%% file: acl_latex.tex
\pdfoutput=1

\documentclass[11pt]{article}

\usepackage[final]{acl}

\usepackage[utf8]{inputenc} 
\usepackage[T1]{fontenc}    
\usepackage{hyperref}       
\usepackage{url}            
\usepackage{booktabs}       
\usepackage{amsfonts}       
\usepackage{amsmath}        
\usepackage{nicefrac}       
\usepackage{microtype}      
\usepackage[table]{xcolor}         
\usepackage{multirow}
\usepackage{tcolorbox}
\usepackage{tabularx}
\usepackage{ragged2e}

\usepackage{enumitem}

\usepackage{graphicx}
\usepackage{longtable}
\usepackage{caption}
\usepackage{geometry}
\usepackage{adjustbox}
\usepackage{float}      

\usepackage{colortbl}  
\usepackage{array}     
\usepackage{amsmath}   
\usepackage{booktabs}  
\usepackage{makecell}  
\usepackage{wrapfig}   

\usepackage{graphicx}
\usepackage{subfigure}
\usepackage{url}

\usepackage{bm}

\usepackage{times}
\usepackage{latexsym}

\usepackage[T1]{fontenc}
\usepackage{pifont}

\usepackage[utf8]{inputenc}
\usepackage{array}
\usepackage{xcolor}

\usepackage[utf8]{inputenc}

\usepackage{microtype}

\usepackage{pifont} 

\usepackage{amssymb}   

\usepackage{inconsolata}

\usepackage{graphicx}

\title{From Coarse to Fine: Benchmarking and Reward Modeling for \\ Writing-Centric Generation Tasks}

\author{
 \textbf{Qingyu Ren\textsuperscript{1}},
 \textbf{Tianjun Pan\textsuperscript{1}},
 \textbf{Xingzhou Chen\textsuperscript{1}},
 \textbf{Xuhong Wang\textsuperscript{2}\thanks{\ Corresponding author.}}
\\
    \textsuperscript{\rm 1}Shanghai Key Laboratory of Data Science,\\ College of Computer Science and Artificial Intelligence, Fudan University,\\
    \textsuperscript{\rm 2}Shanghai Artificial Intelligence Laboratory\\
     \{qyren24,tjpan24, xzchen24\}@m.fudan.edu.cn, wangxuhong@pjlab.org.cn
}

\begin{document}
\maketitle
\begin{abstract}
\input{sections/abstract}

\end{abstract}

\section{Introduction}

\input{sections/intro}

\section{Related Work}

\input{sections/related}

\input{sections/method}

\section{Experiment}

\input{sections/experiment}

\section{Conclusion}
\input{sections/conclusion}

\section*{Limitations}

\input{sections/limitations}

\section*{Ethical Considerations}
\input{sections/ethical}

\section*{Acknowledgments}

\input{sections/ack}




\bibliography{custom}

\appendix
\clearpage
\newpage
\section{Appendix}
\input{sections/appendix}

\end{document}

%% file: sections/abstract.tex
Large language models have achieved remarkable progress in text generation but still struggle with generative writing tasks. In terms of evaluation, existing benchmarks evaluate writing reward models coarsely and fail to measure performance from the perspective of specific requirements. In terms of training, existing training methods either use LLM-as-a-judge approaches or train coarse-grained reward models, lacking fine-grained requirement-adherence reward modeling. To address these issues, we propose a fine-grained evaluation pipeline WEval for writing reward models and a fine-grained reinforcement learning training framework WRL. The evaluation data of WEval covers multiple task categories and requirement types, enabling systematic evaluation of writing reward models by measuring the correlation between the rankings of the reward model and gold rankings. WRL constructs positive and negative samples by selectively dropping instruction requirements, allowing for more precise reward model training. Experiments show that our models achieve substantial improvements across various writing benchmarks and exhibit strong generalization. The code and data are publicly available at \href{https://github.com/Rainier-rq1/From_Coarse_to_Fine}{https://github.com/Rainier-rq1/From\_Coarse\_to\_Fine}.

%% file: sections/intro.tex
Large language models have made significant progress in text generation~\cite{nagano2025llm,sen2025context,wu2025survey}. However, generative writing tasks present significant challenges for these models, such as creative writing~\cite{liao2025rlmr,wei2025igniting,li2026llm}, story generation~\cite{wang2024guiding,venkatraman2025collabstory,liu2026retell}, and report generation~\cite{ding2024fashionregen,wang2025llm,yuan2024continued}. These tasks require models to have high-level comprehension, logical reasoning~\cite{xu2026adaptive,cheng2026empowering}, and  instruction-following capabilities~\cite{bai2024longwriter,wu2025superwriter,ren2025instructions,he2024complex}.

\input{figures/motivation}
\input{figures/framework}

The instructions in writing tasks often contain many specific requirements, and the model’s  output needs to satisfy those requirements. For improving the writing ability of models, current training paradigms can be categorized into supervised fine-tuning and reinforcement learning. Both supervised fine-tuning (SFT) and direct preference optimization (DPO) rely on high-quality outputs or preference pairs, which are difficult to obtain~\cite{pham2024suri,bai2024longwriter,ren2025stepbystepmasteryenhancingsoft}. The reinforcement learning with verifiable rewards (RLVR) training paradigm addresses this problem, requiring only instructions as input and verifiable rewards for reinforcement learning~\cite{wu2025longwriter,wu2025superwriter,chen2025ace}.

Current evaluation and RLVR training paradigms for writing tasks have the following limitations, as shown in Fig.~\ref{fig:motivation}. \textbf{In terms of evaluation}, current benchmarks provide coarse-grained, general evaluation of writing reward models across dimensions like knowledge, reasoning, and safety~\cite{tan2024judgebench,lambert2025rewardbench,huang2026long}, but fail to offer fine-grained assessment of reward modeling performance on specific requirements in writing tasks. \textbf{In terms of training paradigms}, existing methods either provide rewards using a strong reasoning model in the LLM-as-a-judge manner~\cite{chen2025ace,peng2025verif}, which is computationally expensive, potentially biased, and overdependent on the  evaluation LLM’s capability, or train a coarse-grained reward model based on attributes such as fluency, coherence, and helpfulness~\cite{wu2025longwriter,yuan2024self}. These approaches lack fine-grained scoring based on requirement adherence, which limits the effectiveness of RLVR training.

To address these issues, we propose a fine-grained evaluation pipeline WEval for writing reward models and a reinforcement learning training framework WRL  shown in Fig.~\ref{fig:framework}. Our approach applies dropout to requirements in complex writing queries to construct positive and negative examples, where more requirement dropouts result in worse requirement adherence, thereby naturally forming the golden rankings. We evaluate reward models by comparing their rankings with these golden rankings through correlation metrics. Furthermore, the positive and negative examples obtained via requirement dropout enable fine-grained Bradley-Terry (BT) training for writing reward models, which precisely scores model responses based on requirement adherence for effective RLVR training.


In summary, our contributions are as follows: (1) We propose a fine-grained evaluation pipeline WEval for writing reward models; (2) We introduce a fine-grained training framework WRL; (3) We demonstrate that the trained writing reward models effectively enhance the writing capabilities of policy models through reinforcement learning training, and that the writing reward model evaluation results are consistent with the RL training performance across different reward models.

%% file: figures/motivation.tex
\begin{figure}[t] 
    \centering
            \includegraphics[width=0.45\textwidth]{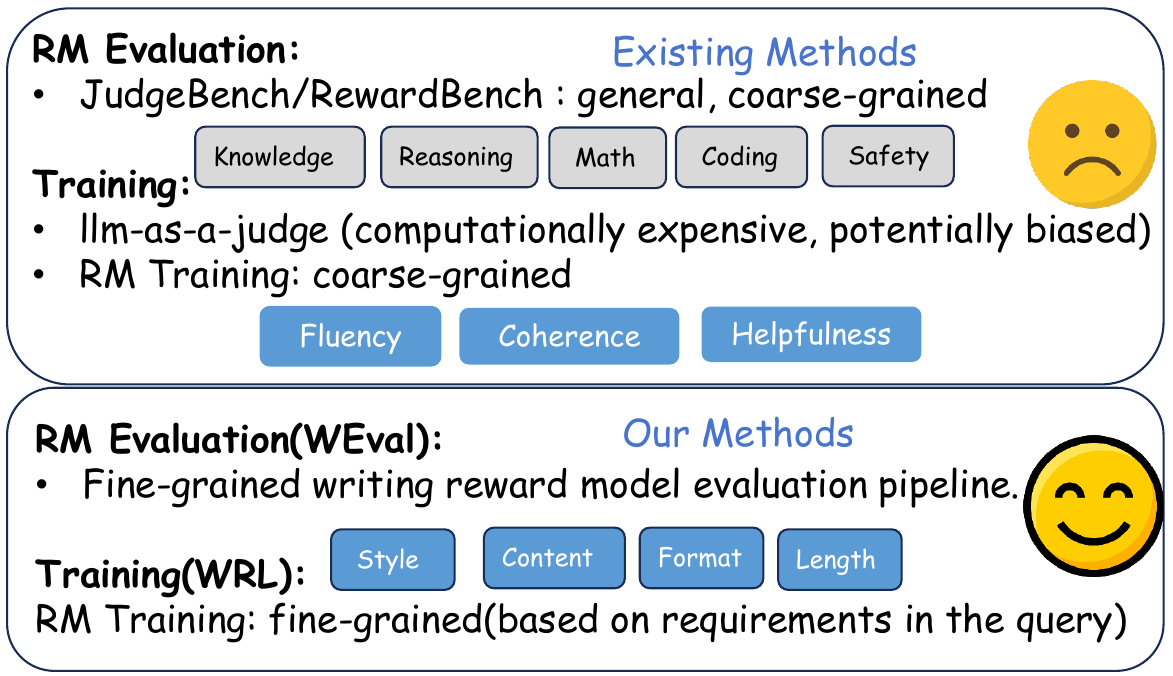}
    \caption{Comparison with prior writing-task evaluation and training paradigms.}
    \label{fig:motivation}
\end{figure}

%% file: figures/framework.tex
\begin{figure*}[t] 
    \centering
            \includegraphics[width=\textwidth]{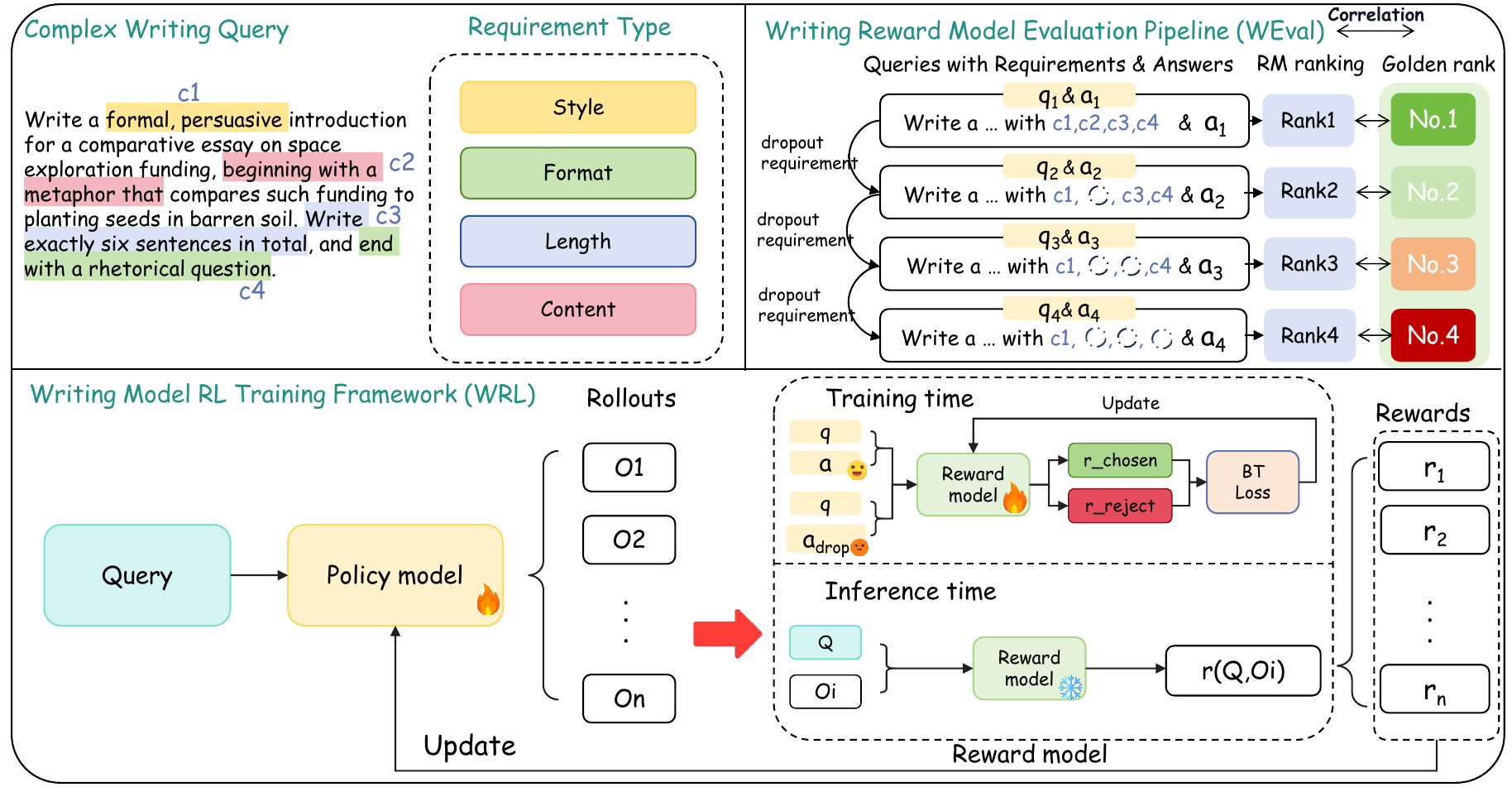}
    \caption{Our framework comprises two components: (WEval), an evaluation pipeline that applies requirement dropout to construct a partial order over requirement adherence, from which the golden rankings are derived, and evaluates reward models by measuring their correlation with the golden rankings; and (WRL), a reinforcement learning framework with fine-grained Bradley-Terry training for reward models, which provide reward signals for RL training.}
    \label{fig:framework}
\end{figure*}

%% file: sections/related.tex
\subsection{Reward Model Evaluation  Benchmarks}

Existing reward model evaluation benchmarks suffer from coarse-grained evaluation dimensions that fail to assess performance on specific requirements. RewardBench~\cite{lambert2025rewardbench} evaluates reward models across general dimensions (Chat, Chat Hard, Safety, Reasoning) but only measures overall accuracy on pairwise comparisons, without decomposing complex instructions into specific requirement types. Similarly, JudgeBench~\cite{tan2024judgebench} evaluates LLM-as-a-Judge systems across dimensions (decision-making ability, consistency, fairness, robustness) but these dimensions are too coarse-grained to systematically evaluate reward models' ability to distinguish between responses with different levels of requirement adherence. Both benchmarks rely on binary pairwise judgments and task-level evaluation, which cannot capture the nuanced ranking quality needed for fine-grained writing evaluation. Different from these benchmarks, our evaluation pipeline enables fine-grained assessment by comparing reward models' rankings of multiple candidate responses with naturally formed golden rankings, systematically evaluating the quality of writing reward models on specific requirements.

\subsection{Training Methods for Writing  Improvement}

Existing training methods for writing models can be divided into supervised fine-tuning and reinforcement learning~\cite{wu2025superwriter,pham2024suri}. SFT training focuses on building high-quality datasets of instruction–response pairs for writing tasks~\cite{quan2024language}. LongWriter~\cite{bai2024longwriter} is trained in stages using SFT and DPO~\cite{rafailov2023direct} to enhance long-text quality. Recently, many researchers have trained writing models using RLVR. ACE-RL~\cite{chen2025ace} employs Qwen3-8B-as-a-judge for constraint verification—which relies heavily on external large models and leads to high computational costs. LongWriter-Zero~\cite{wu2025longwriter} trains its reward model based on coarse-grained attributes such as fluency, coherence, and helpfulness. In contrast, our method trains the reward model in a fine-grained manner by constructing training data according to the writing requirements in the instructions.

%% file: sections/method.tex


\section{Method}
As shown in Fig.~\ref{fig:framework}, our method consists of two main components: (1) a fine-grained evaluation pipeline WEval for writing reward models that applies requirement dropout to construct  a partial order and evaluates reward models by computing the correlation between their rankings of candidate responses and naturally formed golden rankings; and (2) a reinforcement learning training framework WRL that employs fine-grained Bradley--Terry training to writing reward models, which provide reward signals for policy model training via Group Relative Policy Optimization~\cite{shao2024deepseekmath}.

\subsection{Evaluating Writing Reward Model}

\subsubsection{Seed Query Collection}

To construct evaluation data for writing reward models, we collect  seed instructions from large-scale user--LLM conversation datasets. Specifically, we draw from LMSYS-1M~\cite{zheng2023lmsys}, which contains around one million dialogues spanning 25 LLMs collected on the chat platform; WildChat~\cite{zhao2024wildchat}, comprising 652.1k conversations with GPT-3.5 and GPT-4; and PRISM~\cite{kirk2024prism}, a curated set of 8.0k dialogues designed to capture  user preferences and behaviors. 

Our evaluation dataset covers five main task types: (1) Creative Writing \& Narrative, (2) Frameworks \& Structured Plans, (3) Long-form Academic Writing, (4) Discussion \& Expression Tasks, and (5) Informational \& Practical Writing. Given the seed instructions and predefined task categories, we obtain task-aligned subsets through a filtering process. For each category, human annotators first select 10 high-quality prototype examples. Based on their instruction embeddings, we compute a category centroid. For a candidate prompt with embedding $v_j$, its relevance is measured by the cosine similarity:
\[
\text{sim}(v_j, \text{centroid}) = \frac{v_j \cdot \text{centroid}}{\|v_j\| \, \|\text{centroid}\|}.
\]

\subsubsection{Query Variation}

Building on the collected seed instructions, we introduce four major categories of writing requirements: (1) Content, which specifies the key information, themes, or points that the writing task should cover; (2) Style, which defines the tone, language, or stylistic manner expected in the response; (3) Format, which describes the organizational structure or presentation form of the writing task; and (4) Length, which sets constraints on the expected scope, such as the number of words, sentences, or paragraphs.

\subsubsection{Requirement-Level Evaluation}

Existing benchmarks provide coarse-grained, general evaluation of writing reward models but fail to assess performance on specific requirements in writing tasks. To address this gap, we propose a fine-grained evaluation pipeline WEval that evaluates reward models by comparing the correlation between their rankings of candidate responses and naturally formed golden rankings.

For each query $q$ with $n$ requirements ($c_1, c_2, \ldots, c_n$), we generate $n$ candidate responses. Specifically, we perform $n$ rounds: in the first round, we use the original query with all $n$ requirements; in the subsequent $n-1$ rounds, we randomly dropout 1, 2, \ldots, $n-1$ requirements, respectively, and use DeepSeek-R1~\cite{guo2025deepseek} to generate an answer for each modified query. Since more requirement dropouts naturally result in worse requirement adherence, the responses naturally form an ordering where the degree of requirement adherence decreases as more requirements are dropped. This natural ordering serves as the golden ranking. During evaluation, the reward model ranks the $n$ candidate responses $(a_1, a_2, \ldots, a_n)$ for each query $q$. We then assess the reward model's performance by comparing its predicted rankings with the golden rankings through correlation metrics, systematically evaluating the quality of writing reward models on specific requirements. Tab.~\ref{tab:sta} presents the evaluation dataset statistics.

\input{tables/statistic}

Specifically, we use three metrics---\textbf{Correlation}, \textbf{IL}, and \textbf{PL}---to evaluate the ranking produced by the reward model. Let $r_1$ denote the ranking generated by the reward model and $r_2$ denote the golden ranking. The metrics are as follows:

\textbf{Correlation} measures the correlation between the two rankings:

\vspace{0.6em} 

\resizebox{0.9\columnwidth}{!}{%
$\displaystyle
\mathrm{Correlation}
=
\frac{\displaystyle\sum_{1 \le i < j \le n}
\operatorname{sgn}\bigl(r_1(i)-r_1(j)\bigr)\,
\operatorname{sgn}\bigl(r_2(i)-r_2(j)\bigr)}
{\tfrac{1}{2}\,n(n-1)}
$%
}

\vspace{0.6em} 

\textbf{IL (Instruction-Level)} quantifies the proportion of items that occupy the same relative positions:

\begin{equation*}
\mathrm{IL}(r_1,r_2)
=
\frac{1}{n}\,
\bigl|\{\,i \mid \mathrm{pos}_{r_1}(i)=\mathrm{pos}_{r_2}(i)\}\bigr|
\end{equation*}

\textbf{PL (Prompt-Level)} measures the proportion of cases where the two rankings are identical:

\begin{equation*}
\mathrm{PL}(r_1,r_2)
=
\begin{cases}
1,& \text{if } r_1 = r_2\\
0,& \text{otherwise}
\end{cases}
\end{equation*}

\subsection{WRL Training}

\subsubsection{Reward Model Training}

As shown in Fig.~\ref{fig:framework}, our training framework WRL constructs positive and negative examples by \textbf{applying dropout to the requirements} in writing queries, enabling fine-grained reward modeling. We train Qwen2.5-7B-Instruct as the writing reward model. Specifically, for each training sample, given a query $q$ with requirements $(c_1, c_2, \ldots, c_n)$, the original response is treated as a chosen example $a$, while rejected examples $a_{\text{drop}}$ are constructed by dropping requirements from $q$. The reward model is optimized using the Bradley--Terry loss~\cite{bradley1952rank}, encouraging higher scores for responses that better adhere to the requirements:
\begin{equation*}
\mathcal{L}_w = - \log \sigma \big( r(q, a) - r(q, a_{\text{drop}}) \big)
\end{equation*}

\input{tables/rabench}

\subsubsection{GRPO Training}

We collect diverse instructions from WildChat-1M~\cite{zhao2024wildchat} and filter out writing-related tasks using DeepSeek-R1, resulting in a total of 13,221 instructions. We use DeepSeek-R1 to extract all the requirements in the question for reward modeling. During GRPO training, given a query $q$, the policy model generates rollouts $(o_1, o_2, \ldots, o_n)$. For each rollout $o_i$, our trained reward model computes a reward $r(q, o_i)$. These rewards $(r_1, r_2, \ldots, r_n)$ are then used to update the policy model, creating a feedback loop for improving response generation.

%% file: tables/statistic.tex
\begin{table}[t]
\centering

\label{tab:conifer_statistics}
\resizebox{\columnwidth}{!}{%
\begin{tabular}{ccccl}
\toprule
\textbf{\# Inst.} & \textbf{\# Requirements} & \textbf{Candidate Resp.} & \textbf{\# Num} \\
\midrule
\multirow{5}{*}{2750} & \multirow{5}{*}{13750} & $a_1$ & 550 \\
                               &                      & $a_2$ & 550\\
                               &                      & $a_3$ & 550 \\
                               &                      & $a_4$ & 550 \\
                               &                      & $a_5$ & 550 \\
\bottomrule
\end{tabular}%
}
\caption{Statistics of the evaluation dataset of WEval. \# Inst. denotes the number of instructions. \# Requirements denotes the total number of requirements. Candidate Resp. indicates the type of candidate responses generated for each query. \# Num denotes the number of candidate responses of each type.}

\label{tab:sta}
\end{table}

%% file: tables/rabench.tex
\definecolor{lightblue}{RGB}{173, 216, 230}

\begin{table*}[!ht]

\centering

\footnotesize

\resizebox{1.0\textwidth}{!}{

\begin{tabular}{l|c|cccccc|cc|cc|cc}

\toprule

\multirow{2}{*}{Models} & \multirow{2}{*}{\textbf{Avg}} & \multicolumn{6}{c|}{\textbf{Domains}} & \multicolumn{6}{c}{\textbf{Requirements}} \\

 \cmidrule(lr){3-8} \cmidrule(lr){9-14}

& & D1 & D2 & D3 & D4 & D5 & D6 & R1 & C & R2 & C & R3 & C \\ 

\midrule

\rowcolor{lightblue} \multicolumn{14}{l}{\textit{\textbf{Proprietary LLMs}}} \\

\midrule

o3-2025-04-16 & 85.3 & 84.8 & 85.2 & 83.9 & 85.9 & 85.8 & 86.8 & 85.1 & 87.5 & 85.2 & 91.0 & 86.3 & 87.2 \\

Gemini-2.5-pro-preview & 83.1 & 83.2 & 81.5 & 83.0 & 84.5 & 84.5 & 82.1 & 83.6 & 86.5 & 83.9 & 90.5 & 83.4 & 84.0 \\

Claude-3-7-sonnet & 78.5 & 78.2 & 77.9 & 76.5 & 79.4 & 79.3 & 80.9 & 79.4 & 82.5 & 78.8 & 86.1 & 79.2 & 80.5 \\

GPT-4o & 75.5 & 74.4 & 73.4 & 74.4 & 77.9 & 75.9 & 78.1 & 76.8 & 81.6 & 75.8 & 85.5 & 76.1 & 76.7 \\

o1-Preview & 68.6 & 68.5 & 67.0 & 66.6 & 69.5 & 70.3 & 71.4 & 70.1 & 75.1 & 68.5 & 79.8 & 70.9 & 73.8 \\

\midrule

\rowcolor{lightblue} \multicolumn{14}{l}{\textit{\textbf{Open-source LLMs}}} \\

\midrule

DeepSeek-R1-0528 & 83.2 & 83.2 & 81.5 & 81.6 & 85.7 & 84.1 & 84.4 & 84.2 & 87.3 & 83.7 & 89.4 & 83.8 & 82.7 \\

Qwen3-235B-A22B-thinking & 81.5 & 80.2 & 79.2 & 81.0 & 82.9 & 82.5 & 82.9 & 82.5 & 85.0 & 81.3 & 88.2 & 81.3 & 81.8 \\

LongWriter-Zero-32B & 80.3 & 80.7 & 80.3 & 80.2 & 76.1 & 83.6 & 81.0 & 79.9 & 83.4 & 80.8 & 86.8 & 80.2 & 82.1 \\

Qwen3-235B-A22B & 73.6 & 73.6 & 72.9 & 74.0 & 70.1 & 76.5 & 74.7 & 77.5 & 82.1 & 77.0 & 87.3 & 76.3 & 79.6 \\

Qwen2.5-72B-instruct & 65.3 & 65.8 & 63.4 & 63.8 & 62.8 & 68.1 & 67.9 & 65.8 & 70.5 & 65.9 & 78.7 & 66.4 & 68.0 \\

LongWriter-glm4-9B & 62.9 & 64.1 & 63.7 & 62.4 & 61.3 & 65.0 & 61.3 & 62.8 & 66.7 & 63.6 & 74.8 & 63.4 & 65.9 \\

LongWriter-llama3.1-8B & 58.0 & 60.1 & 59.3 & 57.6 & 56.0 & 58.4 & 56.7 & 58.1 & 61.4 & 58.6 & 67.6 & 59.1 & 63.0 \\

Llama-3.3-70B-instruct & 50.4 & 50.7 & 49.3 & 47.9 & 48.5 & 52.9 & 56.6 & 50.7 & 50.7 & 50.4 & 50.4 & 51.1 & 51.1 \\

\midrule
\rowcolor{lightblue} \multicolumn{14}{l}{\textit{\textbf{Capability-enhanced LLMs}}} \\
\midrule

Qwen2.5-1.5B-Instruct & 44.6 & 46.0 & 46.8 & 43.8 & 37.5 & 48.4 & 45.0 & 44.1 & 47.8 & 44.9 & 54.5 & 43.2 & 43.7 \\

\quad\textit{w/ WRL} & \textbf{50.1} & \textbf{53.5} & \textbf{51.7} & \textbf{50.6} & \textbf{42.9} & \textbf{53.9} & \textbf{49.6} & \textbf{49.8} & \textbf{53.2} & \textbf{51.2} & \textbf{61.6} & \textbf{49.7} & \textbf{49.3} \\

\midrule

Qwen2.5-7B-Instruct & 57.0 & 58.6 & 55.9 & 55.3 & 53.0 & 59.5 & 60.1 & 57.4 & 62.3 & 57.7 & 70.2 & 56.6 & 55.6 \\

\quad\textit{w/ WRL} & \textbf{64.4} & \textbf{66.6} & \textbf{63.7} & \textbf{64.9} & \textbf{59.0} & \textbf{69.1} & \textbf{65.7} & \textbf{65.1} & \textbf{68.6} & \textbf{65.8} & \textbf{78.3} & \textbf{64.8} & \textbf{67.3} \\

\midrule

Llama-3.1-8B-Instruct & 47.0 & 48.5 & 46.6 & 44.7 & 43.4 & 50.3 & 51.9 & 46.9 & 50.5 & 47.4 & 56.6 & 46.8 & 45.2 \\

\quad\textit{w/ WRL} & \textbf{48.5} & \textbf{50.4} & \textbf{48.3} & \textbf{46.4} & \textbf{44.4} & \textbf{51.7} & \textbf{52.7} & \textbf{48.2} & \textbf{51.7} & \textbf{49.0} & \textbf{59.0} & \textbf{48.5} & \textbf{47.9} \\

\midrule

Distill-Qwen-14B & 60.7 & 60.0 & 59.7 & 61.7 & 58.0 & 62.5 & 63.9 & 62.1 & 67.4 & 61.7 & 73.1 & 60.7 & 61.2 \\

\quad\textit{w/ WRL} & \textbf{66.7} & \textbf{67.2} & \textbf{64.8} & \textbf{67.2} & \textbf{64.3} & \textbf{69.0} & \textbf{69.8} & \textbf{68.3} & \textbf{72.7} & \textbf{67.8} & \textbf{80.8} & \textbf{67.5} & \textbf{71.0} \\

\midrule

Qwen3-8B & 74.9 & 74.9 & 73.9 & 75.0 & 73.5 & 76.8 & 76.5 & 75.6 & 80.1 & 75.7 & 84.8 & 74.6 & 78.1 \\

\quad\textit{w/ WRL} & \textbf{76.2} & \textbf{76.3} & \textbf{74.6} & \textbf{76.5} & \textbf{75.2} & \textbf{78.3} & \textbf{78.0} & \textbf{77.6} & \textbf{80.7} & \textbf{76.9} & \textbf{86.6} & \textbf{77.0} & \textbf{81.3} \\

\bottomrule

\end{tabular}}

\caption{Performance of different LLMs on WritingBench across six domains and three writing requirements. Scores are normalized from a 0--10 range to a 100-point scale. The domains include: (D1) Academic \& Engineering, (D2) Finance \& Business, (D3) Politics \& Law, (D4) Literature \& Art, (D5) Education, and (D6) Advertising \& Marketing. The writing requirements assessed are: (R1) Style, (R2) Format, and (R3) Length. Here, ``C'' indicates category-specific scores. The latest results are available on the online leaderboard.}

\label{tab:main table}

\end{table*}

%% file: sections/experiment.tex
\input{tables/longwriter}

\subsection{Experiment Setup}

We evaluate our method on three categories of models: (1) \textbf{Proprietary LLMs}: o3-mini, Gemini-2.5-Pro-Preview, Claude-3-7-Sonnet, GPT-4o, and o1-Preview; (2) \textbf{Open-source LLMs}: DeepSeek-R1-0528, Qwen3 series~\cite{yang2025qwen3}, Qwen2.5 series, and Llama-3 series~\cite{grattafiori2024llama}; (3) \textbf{Writing-enhanced LLMs}: LongWriter-llama3.1-8B, LongWriter-glm4-9B~\cite{bai2024longwriter}, and LongWriter-Zero-32B~\cite{wu2025longwriter}. We apply WRL to base models including Qwen2.5 series, Llama-3.1-8B-Instruct, Distill-Qwen-14B, and Qwen3-8B.

We evaluate writing models on multiple benchmarks to assess both in-domain and out-of-domain performance. For \textbf{in-domain evaluation}, we use WritingBench~\cite{wu2025writingbench}, LongWriter~\cite{bai2024longwriter}, and Arena-Write~\cite{wu2025longwriter}. For \textbf{out-of-domain evaluation}, we test on DeepResearch Bench-RACE~\cite{du2025deepresearch} and FINDER\_DEFT~\cite{zhang2025far}. For \textbf{reward model evaluation}, we evaluate on our constructed evaluation dataset.

\input{figures/arena}
\subsection{Performance of Writing Models}
\input{figures/dp}


As reported in Tab.~\ref{tab:main table}, WRL consistently improves the performance of base models across different parameter scales on WritingBench. For example, Qwen2.5-7B-Instruct improves by 7.4 and outperforms writing-oriented models such as LongWriter-llama3.1-8B and LongWriter-glm4-9B with comparable model sizes. Qwen2.5-1.5B-Instruct improves by 5.5, while Distill-Qwen-14B and Qwen3-8B achieve gains of 6.0 and 1.3, respectively. Notably, Qwen3-8B enhanced with WRL even surpasses proprietary models such as GPT-4o and o1-Preview. These gains are observed across all six domains and three writing requirements, suggesting that our fine-grained reward modeling strategy generalizes well across different writing scenarios.

Tab.~\ref{tab:longwriter} further shows that, after training with our method, the model achieves markedly better writing quality across multiple evaluation dimensions, including Relevance, Accuracy, Coherence, Clarity, Breadth and Depth, and Reading Experience, demonstrating the practical value of our method in real-world writing applications.

Similar trends can be observed in Fig.~\ref{fig:as}, where models trained with WRL achieve higher win rates against strong baselines on Arena-Write. Qwen2.5-7B-Instruct-WRL improves the win rate by 4.6\%, and Distill-Qwen-14B-WRL achieves the largest gain of 8.7\%, further indicating better alignment with human preferences.

\subsection{Performance of Writing Reward Models}

As shown in Tab.~\ref{tab:resultsff}, we evaluate a diverse set of reward models on our evaluation dataset, including both LLM-as-a-judge methods and trained reward models. Compared with strong baselines, our 7B reward model achieves the best performance consistently across all three metrics. Specifically, it obtains 94.6 on Correlation, 97.3 on IL, and 78.0 on PL, outperforming all competing reward models by clear margins. These results demonstrate that our reward model is more aligned with fine-grained writing quality assessment and provides more reliable reward signals than both judge-style prompting methods and existing reward modeling baselines. In addition, the rankings of human annotators achieve strong performance on all three metrics, further validating that our evaluation protocol is highly consistent with human judgments.

\subsection{Ablation Studies}

As shown in Tab.~\ref{tab:abla}, we study the impact of different training strategies and reward models on downstream performance. Our WRL method achieves the best results on both WritingBench and Arena-Write, consistently outperforming all other approaches, including supervised fine-tuning, LLM-as-a-judge RL, and WRL with existing reward models. Notably, replacing the reward model with Our-RM-7B further boosts performance to 64.4 on WritingBench and 26.05 on Arena-Write, yielding the strongest overall results. Moreover, the relative ranking of reward models in downstream RL is consistent with their evaluation results in Tab.~\ref{tab:resultsff}.  These results further validate the effectiveness of our reward model evaluation protocol and show that it is highly predictive of reward model utility in practical RL settings.

\input{tables/rm}
\input{tables/ablation}

\subsection{Generalizability}

Our WRL method generalizes well to DeepResearch tasks and consistently improves the quality of generated research reports. As shown in Fig.~\ref{fig:dp}, WRL yields stable gains on both DeepResearch Bench-RACE and FINDER\_DEFT across models of different scales. On DeepResearch Bench-RACE, Qwen2.5-1.5B-Instruct-WRL improves the overall score by 4.3, while Distill-Qwen-14B-WRL achieves the largest gain of 4.5. On FINDER\_DEFT, WRL also substantially improves checklist pass rate, with gains of 14.3 for Qwen2.5-1.5B-Instruct, 12.1 for Distill-Qwen-14B, and 10.0 for Llama-3.1-8B-Instruct. These results show that WRL transfers effectively to out-of-domain research report generation tasks.

To further examine the generalization of our evaluation framework WEval, we reconstruct the evaluation datasets using two different teacher models, GPT-4o and Gemini-2.5-Pro, and report the results in Tab.~\ref{tab:results1} and Tab.~\ref{tab:results2}, respectively. We observe that the overall ranking of reward models remains highly consistent with that in Tab.~\ref{tab:resultsff} across both settings. In particular, Our-RM-7B consistently achieves the best performance on all three metrics, while Qwen2.5-7B-Instruct-as-a-judge remains the strongest baseline among the compared methods. This consistency across datasets constructed by different models indicates that our evaluation results are robust to the choice of data construction model. These findings further demonstrate the strong generalization ability of our evaluation method and validate its reliability for reward model assessment.

\input{tables/rm1}
\input{tables/rm2}
\input{tables/comp-zero}
\input{figures/comp-pre}
We further verify the generalization of our reward model training approach on Qwen3-32B-Base by comparing with Writing-Zero~\cite{jia2025writing} on RewardBench~\cite{lambert2025rewardbench}. Writing-Zero trains a reward model through a multi-stage pipeline involving preference filtering, cold-start supervised fine-tuning with reasoning traces, and GRPO-based reinforcement learning. In contrast, our method directly constructs preference pairs from dropped requirements and the natural partial order implied by instructions, and optimizes the reward model with the Bradley--Terry loss. As shown in Tab.~\ref{tab:zero}, our trained model substantially improves over the Qwen3-32B-Base baseline and achieves strong overall performance, reaching 97.8 on Chat, 85.1 on Safety, 93.4 on Reasoning, and 85.3 on average. Although it slightly underperforms Pairwise GenRM on Chat Hard, it approaches the performance of Writing-Zero's model with a much simpler training pipeline. These results further demonstrate that our method generalizes effectively to larger base models and provides a simple yet strong solution for reward model training.

\subsection{Analysis}

Fig.~\ref{fig:case} presents a representative case study comparing existing writing evaluation paradigms with our method. Although previous methods may regard Responses A and B as comparable under coarse-grained criteria such as fluency, coherence, and helpfulness, this assessment fails to capture whether the responses truly satisfy the detailed task requirements. In this example, Response A violates the length constraint, while Response B misses multiple explicit requirements, including the rhetorical question, the numbered bullet format, the length requirement, and the required ending sentence. Our method explicitly decomposes the writing task into fine-grained requirement-level criteria and evaluates each response against these concrete requirements. As a result, it can accurately identify that Response A satisfies more key requirements than Response B, leading to a more reliable preference judgment. This case demonstrates that, compared with conventional holistic evaluation paradigms, our method provides a more fine-grained, requirement-aware, and accurate assessment for complex writing tasks.

%% file: tables/longwriter.tex
\begin{table*}[t]

\centering

\setlength{\tabcolsep}{4pt}

\resizebox{\textwidth}{!}{

\begin{tabular}{lccccccc}

\toprule

\textbf{Model} & \textbf{Relevance} & \textbf{Accuracy} & \textbf{Coherence} & \textbf{Clarity} & \textbf{Breadth and Depth} & \textbf{Reading Experience} & \textbf{Total} \\

\midrule

LongWriter-glm4-9B & 95.0 & 90.0 & 91.7 & 88.5 & 73.3 & 83.5 & 87.0 \\

LongWriter-llama3.1-8B & 91.5 & 88.3 & 86.0 & 85.6 & 66.7 & 76.5 & 82.4 \\

\midrule

Qwen2.5-1.5B-Instruct & 68.5 & 71.0 & 65.2 & 65.0 & 42.1 & 55.0 & 61.1 \\
\rowcolor{green!10}
Qwen2.5-1.5B-Instruct-WRL & \textbf{76.9} & \textbf{77.7} & \textbf{70.2} & \textbf{70.0} & \textbf{50.4} & \textbf{57.5} & \textbf{67.1} \\

\midrule

Llama-3.1-8B-Instruct & 89.8 & 81.9 & 80.6 & 79.2 & 53.3 & 66.9 & 75.3 \\
\rowcolor{green!10}
Llama-3.1-8B-Instruct-WRL & \textbf{92.3} & \textbf{84.0} & \textbf{81.9} & \textbf{79.8} & \textbf{55.2} & \textbf{69.8} & \textbf{77.2} \\

\midrule

Qwen2.5-7B-Instruct & 91.7 & 89.6 & 80.8 & 80.6 & 67.7 & 74.6 & 80.8 \\
\rowcolor{green!10}
Qwen2.5-7B-Instruct-WRL & \textbf{92.9} & \textbf{90.0} & \textbf{82.7} & \textbf{82.9} & \textbf{72.3} & \textbf{76.5} & \textbf{82.9} \\

\midrule

Distill-Qwen-14B & 97.1 & \textbf{92.9} & \textbf{94.4} & \textbf{92.1} & 70.0 & 84.0 & 88.4 \\
\rowcolor{green!10}
Distill-Qwen-14B-WRL & \textbf{97.5} & 91.9 & 92.7 & 90.6 & \textbf{76.3} & \textbf{85.2} & \textbf{89.0} \\

\midrule

Qwen3-8B & \textbf{96.2} & 93.3 & 90.2 & 89.6 & 79.2 & \textbf{86.3} & 89.1 \\
\rowcolor{green!10}
Qwen3-8B-WRL & 95.6 & \textbf{93.5} & \textbf{91.3} & \textbf{89.8} & \textbf{81.9} & 85.4 & \textbf{89.6} \\

\bottomrule

\end{tabular}

}

\caption{Performance  of different models on LongWriter. The results show that our method can significantly improve the writing quality of the model.
}

\label{tab:longwriter}

\end{table*}

%% file: figures/arena.tex
\begin{figure}[t] 
    \centering
            \includegraphics[width=0.5\textwidth]{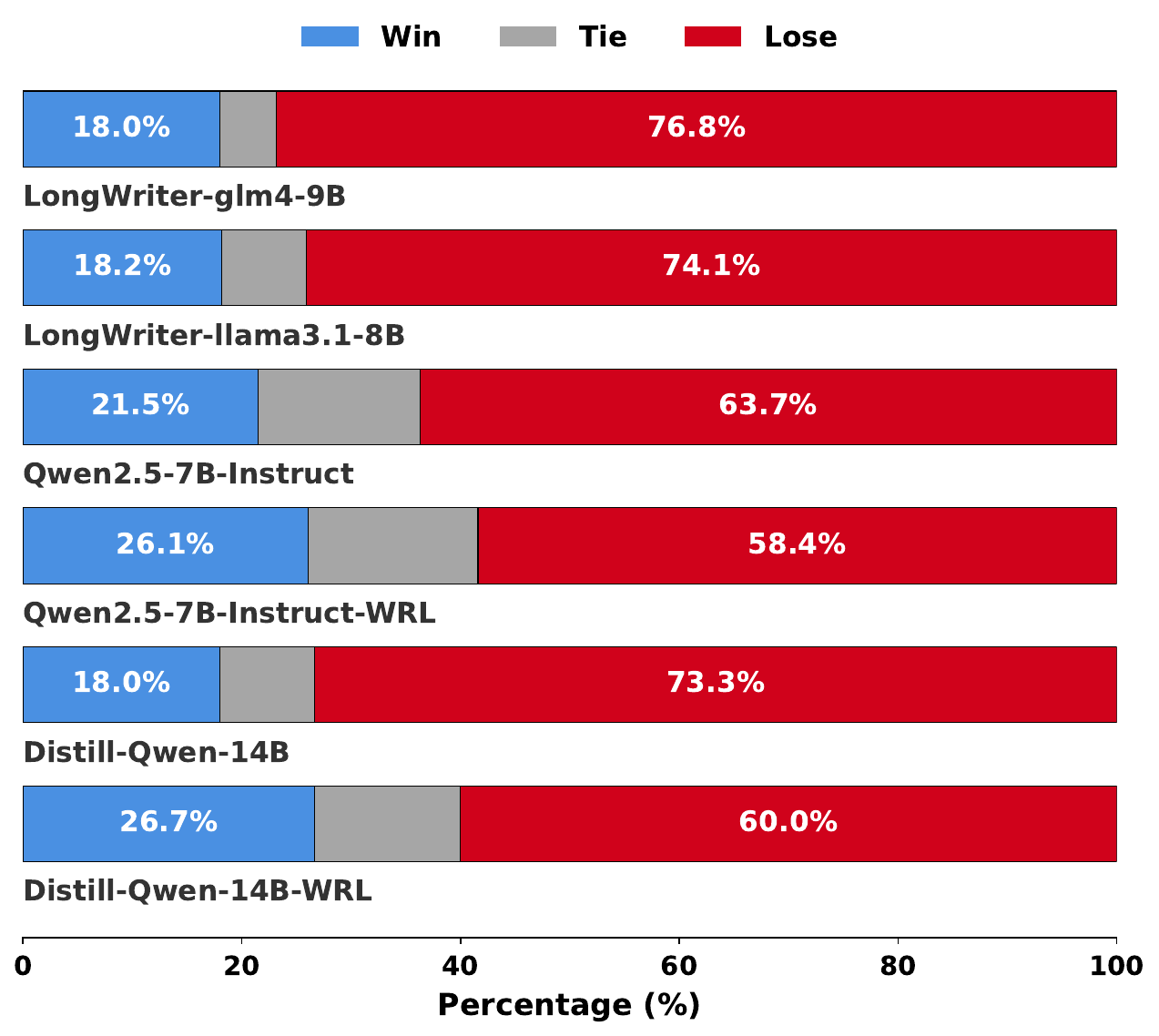}
    \caption{Performance of different models on Arena-
Write against six strong baselines.}
    \label{fig:as}
\end{figure}

%% file: figures/dp.tex
\begin{figure*}[t]
    \includegraphics[width=0.5\textwidth]{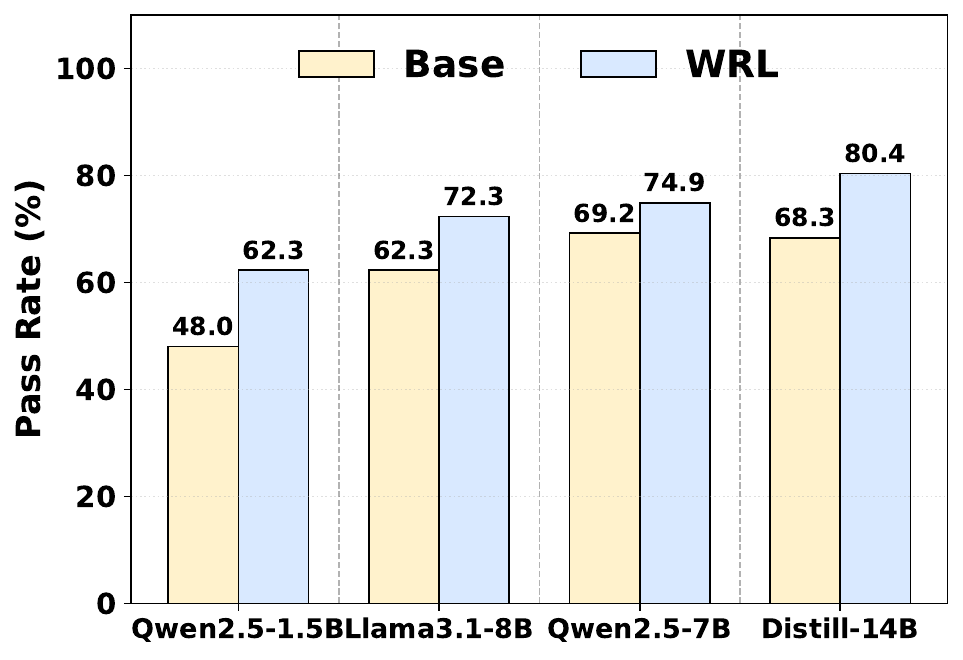}
    \includegraphics[width=0.5\textwidth]{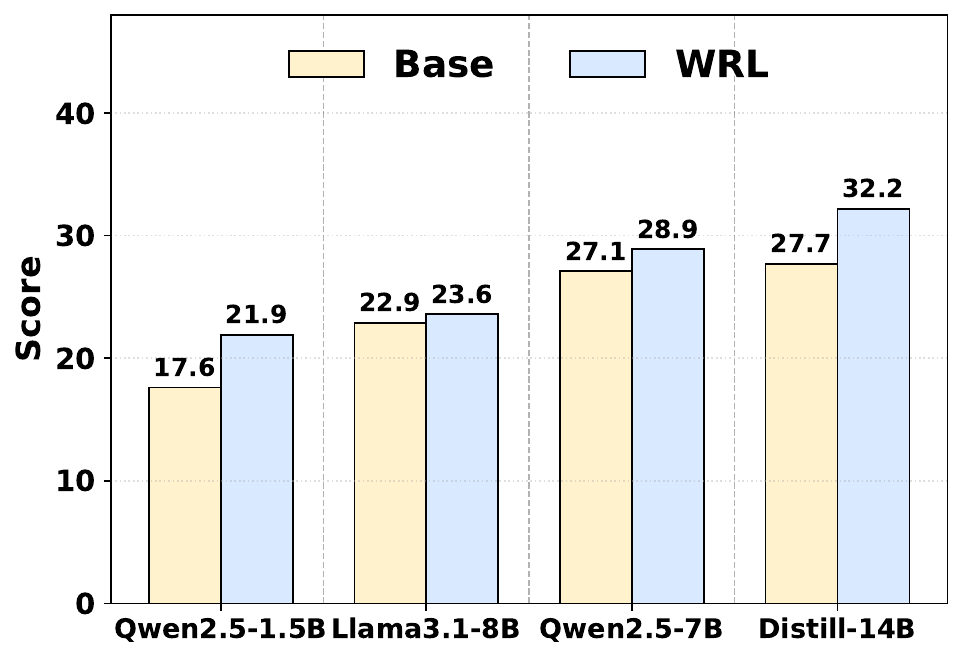}

    \caption{Performance of different models on DeepResearch-related benchmarks, including  FINDER\_DEFT (left) and DeepResearch Bench-RACE (right).}
    \label{fig:dp}
\end{figure*}

%% file: tables/rm.tex
\begin{table}[t]

\centering

\resizebox{\columnwidth}{!}{

\begin{tabular}{lccc}

\toprule

\textbf{Models} & \textbf{Correlation} & \textbf{IL} & \textbf{PL} \\

\midrule

Qwen2.5-7B-Instruct-as-a-judge & 85.4 & 92.9 & 60.9 \\

Qwen2.5-72B-Instruct-as-a-judge & 93.6 & 96.8 & \textbf{78.7} \\


Writing-Critic-7B & 80.6 & 90.8 & 42.7 \\

Skywork-Reward-V2-Llama-3.1-8B & 75.4 & 88.7 & 32.9 \\

Skywork-Reward-V2-Qwen3-8B & 82.5 & 91.7 & 45.3 \\

\rowcolor{green!10}
Our-RM-7B & \textbf{94.6} & \textbf{97.3} & 78.0 \\

\midrule

Human  & 95.7 & 97.9 & 79.4 \\

\bottomrule

\end{tabular}

}

\caption{Performance of different reward models on the  evaluation dataset. Our trained reward model demonstrates superior performance in fine-grained requirement adherence evaluation.}

\label{tab:resultsff}

\end{table}

%% file: tables/ablation.tex
\begin{table}[t]

\centering

\setlength{\tabcolsep}{4pt}

\resizebox{\columnwidth}{!}{

\begin{tabular}{lcccc}

\toprule

\textbf{Models} & \textbf{Reward Models} & \textbf{WritingBench}&\textbf{Arena-Write} \\

\midrule

Qwen-2.5-7B-Inst. & - & 57.0&21.51 \\

w/ SFT & - & 59.4&16.47\\

w/ llm-as-Judge RL & Qwen2.5-7B-Inst. & 62.1&24.70 \\

w/ WRL & Skywork-Qwen3 & 61.2 &23.53 \\
\rowcolor{green!10}

w/ WRL & Our-RM-7B & \textbf{64.4}&\textbf{26.05} \\

\bottomrule

\end{tabular}

}

\caption{Ablation study on different training methods and reward models across WritingBench and Arena-Write benchmarks.}

\label{tab:abla}

\end{table}

%% file: tables/rm1.tex
\begin{table}[t]
\centering
\resizebox{\columnwidth}{!}{
\begin{tabular}{lccc}
\toprule
\textbf{Models} & \textbf{Correlation} & \textbf{IL} & \textbf{PL} \\
\midrule

Qwen2.5-7B-Instruct-as-a-judge & 83.5 & 92.0 & 59.6 \\
Writing-Critic-7B & 80.6 & 90.8 & 43.5 \\
Skywork-Reward-V2-Llama-3.1-8B & 82.8 & 91.9 & 44.0 \\

\rowcolor{green!10}
Our-RM-7B & \textbf{92.5} & \textbf{96.5} & \textbf{76.0} \\
\bottomrule
\end{tabular}
}
\caption{Performance of different reward models on the evaluation dataset constructed by GPT-4o.}
\label{tab:results1}
\end{table}

%% file: tables/rm2.tex
\begin{table}[t]
\centering
\resizebox{\columnwidth}{!}{
\begin{tabular}{lccc}
\toprule
\textbf{Models} & \textbf{Correlation} & \textbf{IL} & \textbf{PL} \\
\midrule

Qwen2.5-7B-Instruct-as-a-judge & 84.2 & 92.4 & 59.6 \\
Writing-Critic-7B & 80.6 & 90.8 & 41.5 \\
Skywork-Reward-V2-Llama-3.1-8B & 79.5 & 90.6 & 43.5 \\

\rowcolor{green!10}
Our-RM-7B & \textbf{91.2} & \textbf{95.8} & \textbf{68.4} \\
\bottomrule
\end{tabular}
}
\caption{Performance of different reward models on the evaluation dataset constructed by Gemini-2.5-Pro.}
\label{tab:results2}
\end{table}

%% file: tables/comp-zero.tex
\begin{table}[t]

\setlength{\tabcolsep}{6pt}

\resizebox{0.5\textwidth}{!}{
\begin{tabular}{lccccc}
\toprule
\textbf{Model} & \textbf{Chat} & \textbf{Chat Hard} & \textbf{Safety} & \textbf{Reasoning} & \textbf{Avg} \\
\midrule
Qwen-32B-Base & 93.9 & 67.3 & 87.0 & 84.3 & 83.1 \\
Claude-3.5-Sonnet        & 96.4 & 74.0 & 81.6 & 84.7 & 84.2 \\
Pairwise GenRM*  & --   & --   & --   & --   & \textbf{87.4} \\
\rowcolor{green!10}
Our Trained RM           & 97.8 & 64.9 & 85.1 & 93.4 & \underline{85.3} \\
\bottomrule
\end{tabular}
}

\vspace{2pt}
\caption{Comparison of different reward models on RewardBench. * denotes that the results are taken directly from the original paper Writing-Zero. We use \textbf{bold} for the best results and \underline{underlined} for the second-best results. Our reward model training method is simple yet effective.
}
\label{tab:zero}
\end{table}

%% file: figures/comp-pre.tex
\begin{figure*}[t]
    \centering
    \includegraphics[width=\textwidth]{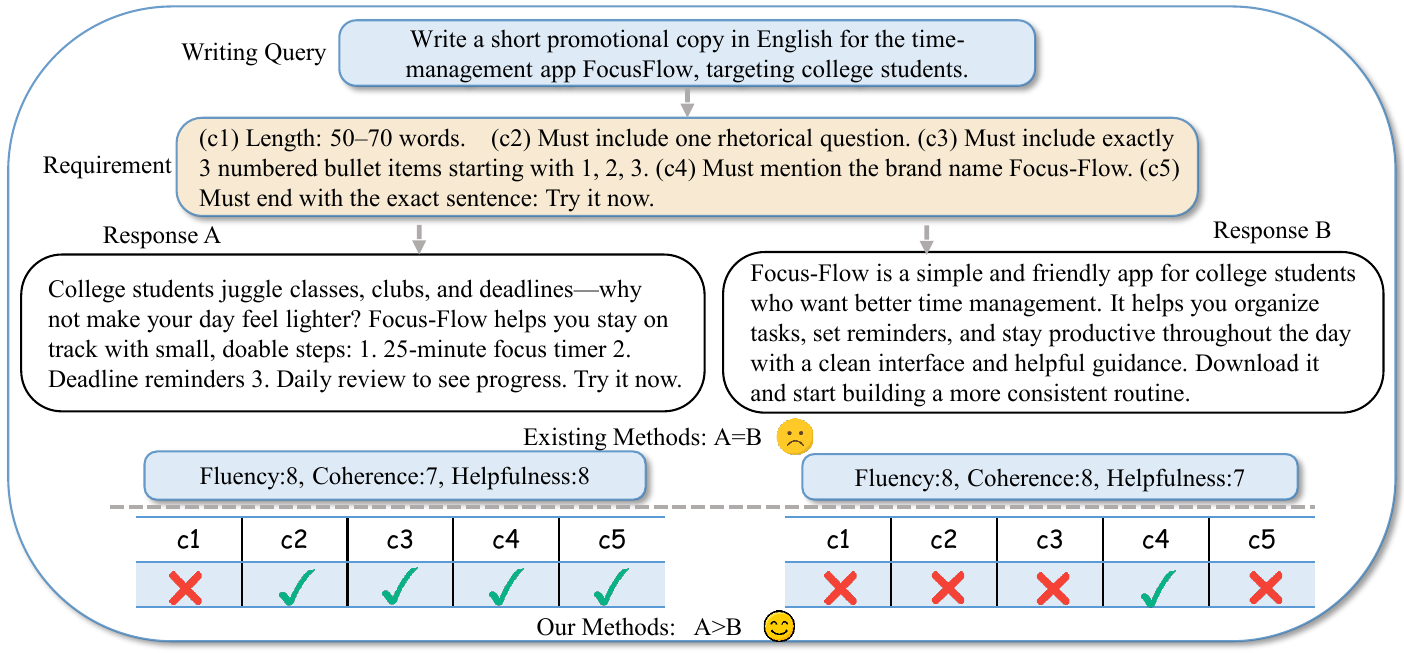}
    \caption{Case study between previous writing task evaluation paradigms and our method. Our approach provides fine-grained, requirement-level evaluation and reward modeling.}
    \label{fig:case}
\end{figure*}

%% file: sections/conclusion.tex
In this paper, we first propose a fine-grained evaluation pipeline WEval for writing reward models that assesses their performance on specific writing requirements through correlation metrics with naturally formed golden rankings constructed via requirement dropout. We introduce a reinforcement learning training framework, WRL, that employs fine-grained Bradley--Terry training to train writing reward models, which are then used to provide reward signals for RL training. Experimental results demonstrate that our trained writing reward models effectively improve policy models' writing capabilities, and the consistency between reward model evaluation results and downstream RL training performance validates the effectiveness of our evaluation approach.

%% file: sections/limitations.tex
Our study has several limitations. First, we do not evaluate our training method on larger-scale models (e.g., 32B parameters or above). Nevertheless, our approach is model-agnostic and shows strong generalization across different architectures and training settings. Second, the constructed evaluation datasets cover only a limited range of requirement types. Although they already capture several representative writing constraints, they do not yet fully reflect the diversity and complexity of real-world writing tasks. 

%% file: sections/ethical.tex
We discuss the potential ethical concerns as follows. The annotation of instruction task categories and the ranking of the reward model evaluation dataset were conducted by three annotators with computer science backgrounds recruited by our institution. The annotation team remained anonymous to the authors. We ensured that the privacy rights of all annotators were respected throughout the annotation process. All annotators were compensated above the local minimum wage and gave informed consent for the data to be used for research purposes. Disagreements in annotation were resolved through majority voting. The annotation details are
shown in Appx.~\ref{appx:ct}.

%% file: sections/ack.tex
We acknowledge the use of \href{https://github.com/cursor/cursor}{Cursor}
 as an AI-assisted writing tool during the preparation of this manuscript. Its role was  to polish the language of the initial draft. All core ideas presented in the paper were conceived independently by the authors.

%% file: sections/appendix.tex
\input{tables/constraint_prompt}

\subsection{Prompt for Evaluation Dataset Construction}
To construct our evaluation dataset, we employ a systematic prompt engineering approach that generates diverse and challenging writing tasks. As shown in Tab.~\ref{tab:constraint_prompt}, we use a comprehensive prompt template that guides the generation of atomic constraints to enhance the difficulty of seed questions. The prompt consists of two main components: (1) \textbf{Task Description}, which specifies the requirements for designing five new atomic constraints that significantly increase the difficulty of answering a given seed question, and (2) \textbf{Constraint References}, which provides eight categories of constraint types including length, format, style and content constraints. Each category contains multiple sub-types with specific examples, enabling the generation of diverse and granular constraints that cover various aspects of writing tasks. The constraints must be specific, actionable, and non-conflicting, ensuring that the resulting evaluation dataset contains high-quality and challenging writing instructions.

\subsection{Details of Tasks in the Evaluation Dataset}
\subsection*{1. Creative Writing \& Narrative}
\textbf{Scope}: Stories, songs, and scripts emphasizing imagination, narrative diversity, and emotional resonance.
\textbf{Purpose}: Engage readers through plot, character development, dialogue, and creative expression.
\textbf{Techniques}: Metaphors, similes, plot twists, rhythm, rhyme, suspense, and perspective shifts.
\textbf{Audience}: Children, young adults, or mature audiences.
\textbf{Formats}: Short stories, musicals, film scripts, serialized narratives, lyrical compositions.

\subsection*{2. Frameworks \& Structured Plans}
\textbf{Scope}: Outlines, conceptual frameworks, structured workflows, and planning templates.
\textbf{Purpose}: Organize ideas, guide projects, and present logical sequences for execution.
\textbf{Techniques}: Hierarchical structures, bullet points, flowcharts, stepwise reasoning, and modular design.
\textbf{Audience}: Teams, students, or professionals requiring clarity and actionable plans.
\textbf{Formats}: Project plans, design frameworks, step-by-step guides, and workflow diagrams.

\subsection*{3. Long-form Academic Writing}
\textbf{Scope}: Extended essays, research papers, technical reports, and scholarly articles integrating citations and empirical data.
\textbf{Purpose}: Demonstrate analysis, argumentation, and evidence-based reasoning.
\textbf{Techniques}: Formal tone, structured headings, literature reviews, methodology sections, and in-text citations.
\textbf{Audience}: Academics, researchers, and specialists in a given discipline.
\textbf{Formats}: Thesis, journal articles, white papers, systematic reviews, and technical manuals.

\subsection*{4. Discussion \& Expression Tasks}
\textbf{Scope}: Question–answer formats, interviews, debates, or dialogue-driven prompts emphasizing reasoning and perspective exchange.
\textbf{Purpose}: Facilitate critical thinking, reflective responses, and conversational clarity.
\textbf{Techniques}: Probing questions, counterarguments, structured dialogues, role-playing, and argument scaffolding.
\textbf{Audience}: Classroom settings, workshops, research interviews, or collaborative problem solving.
\textbf{Formats}: Interviews, Socratic dialogues, debate transcripts, discussion boards, reflective essays.

\subsection*{5. Informational \& Practical Writing}
\textbf{Scope}: Functional, utilitarian writing such as reports, letters, instructions, and procedural documentation.
\textbf{Purpose}: Communicate facts, procedures, or practical guidance clearly and efficiently.
\textbf{Techniques}: Concise language, structured formatting, headings, numbered steps, tables, and bullet points.
\textbf{Audience}: General public, business professionals, students, or technical users.
\textbf{Formats}: Business reports, user manuals, formal letters, policy documents, and how-to guides.

\subsection{Details of Requirements  in the Evaluation Dataset}
\subsection*{1. Length Requirements}
\textbf{Word/Sentence/Paragraph Limits}: Total word count (e.g., 120--150 words), exact sentence count (e.g., 5), paragraph limits (e.g., 3 paragraphs, each $\leq$40 words).
\textbf{Proportions and Lengths}: Proportional allocation (e.g., conclusion 30\%), character restrictions ($\leq$500 including spaces), sentence length limits (e.g., each sentence $<$15 words or $\geq$10 words).

\subsection*{2. Format Requirements}
\textbf{Text Structure}: Custom paragraphs, headings, emphasis, examples, bullet points.
\textbf{Professional Standards}: Specific application formats (e.g., electronic medical records).
\textbf{Emphasis Rules}: Text highlighting methods (e.g., bold key terms, use "!" for warnings).

\subsection*{3. Style Requirements}
\textbf{Tone and Voice}: Humorous, formal, or perspective-based (e.g., historian, retired elder).
\textbf{Rhetorical Devices}: Metaphors, parallelism, rhetorical questions, etc.
\textbf{Audience and Cultural Style}: Adjust for audience, culture, or era (e.g., elementary students, Tang dynasty poet).
\textbf{Emotional Appeal}: Evoke empathy, urgency, or other emotions.

\subsection*{4. Content Requirements}
\textbf{Entities and Data}: Include specific entities (e.g., scientists) and data points.
\textbf{Chronology and Themes}: Sequential events and multiple aspects (economic, environmental, social).
\textbf{Perspective and Interdisciplinary Focus}: Unique time-space viewpoints or combined disciplines.
\textbf{Counterarguments and Hypotheticals}: Address opposing views or create hypothetical scenarios.

\subsection{Evaluation Benchmarks}

\subsubsection{In-Domain Benchmarks:}

\textbf{WritingBench}~\cite{wu2025writingbench}: A comprehensive benchmark comprising 1,239 tasks designed to evaluate LLMs across 6 core writing domains and 100 subdomains, encompassing creative, persuasive, informative, and technical writing. WritingBench supports evaluation of multiple requirement types including length, format, and style constraints, making it suitable for assessing models' ability to follow diverse writing instructions.

\textbf{LongWriter}~\cite{bai2024longwriter}: A benchmark suite consisting of two components: LongBench-Write and LongWrite-Ruler. LongBench-Write focuses on measuring long-form output quality across multiple dimensions including relevance, accuracy, coherence, clarity, breadth and depth, and reading experience, as well as output length. LongWrite-Ruler is designed as a lightweight stress test to evaluate the model's maximum output length capability, measuring how many words a model can generate in a single response.

\textbf{Arena-Write}~\cite{wu2025longwriter}: A small-scale benchmark of 100 user writing tasks collected from real-world scenarios, designed to evaluate long-form generation models in realistic settings. Each task covers diverse formats such as social posts, essays, and reports, with many requiring outputs over 2,000 words. The benchmark uses human preference judgments to assess model performance through win rate comparisons against strong baselines.

\subsubsection{Out-of-Domain Benchmarks:}

\textbf{DeepResearch Bench-RACE}~\cite{du2025deepresearch}: A comprehensive benchmark for evaluating deep research agents' ability to generate high-quality research reports. The benchmark assesses multiple dimensions including comprehensiveness (coverage of research topics), depth (level of detail and analysis), instruction following (adherence to specific requirements), readability (clarity and organization), and overall quality. It tests models' generalization capabilities to research-oriented writing tasks that require extensive knowledge integration and structured output.

\textbf{FINDER\_DEFT}~\cite{zhang2025far}: A benchmark designed to evaluate models' ability to follow detailed formatting and content requirements in research report generation. The benchmark uses a checklist-based evaluation approach, measuring the pass rate of generated reports against a comprehensive set of formatting and content criteria. It focuses on assessing models' precision in adhering to specific structural and content requirements, making it particularly suitable for evaluating fine-grained instruction following capabilities.

\subsection{Baselines}

\textbf{LongWriter-llama3.1-8B}~\cite{bai2024longwriter}: A writing-specialized model based on Meta-Llama-3.1-8B, trained with SFT and DPO. Capable of generating 10,000+ words.

\textbf{LongWriter-glm4-9B}~\cite{bai2024longwriter}: A writing-enhanced model based on GLM-4-9B, trained with SFT and DPO.

\textbf{LongWriter-Zero-32B}~\cite{wu2025longwriter}: A purely RL-based model trained with coarse-grained reward attributes (fluency, coherence, helpfulness).

\subsection{Evaluation Prompt Arena-Write}

\input{tables/eval}

As shown in Tab.~\ref{tab:eval}, we provide the prompts used for the Arena-Write evaluation.





\subsection{RL Training Implementation Details}

We conduct training using the VeRL framework with the GRPO algorithm on the subset of \textit{WildChat} dataset. Prompts and responses are truncated to a maximum length of 12000 tokens each. Data are shuffled with a fixed random seed of 1. Rollouts are generated with a rollout batch size of 384. Training is performed on a single node with 8 H200 GPUs. The global batch size is set to 96, with micro-batches of size 2 per device for policy updates and micro-batches of size 8 per device for experience generation. Rollouts are sampled with a temperature of 1.0 and a group size of 5, using tensor parallelism of size 2 and a maximum of 25{,}000 batched tokens.

\subsection{Annotation Details}
\label{appx:ct}
Annotators were asked to complete two annotation tasks. We present the \textit{Instructions Given To Participants} as follows:

\paragraph{Task 1: Instruction category annotation.}
For each writing instruction, assign it to the single most appropriate task category based on its main purpose and form. Use the following five categories defined in the paper: (1) Creative Writing \& Narrative, (2) Frameworks \& Structured Plans, (3) Long-form Academic Writing, (4) Discussion \& Expression Tasks, and (5) Informational \& Practical Writing. Choose the category that best reflects the primary writing objective of the instruction.

\paragraph{Task 2: Response ranking based on requirement adherence.}
For each query, you will be shown multiple candidate responses. Rank these responses according to how well they satisfy the requirements in the original writing prompt. A response should receive a higher rank if it follows more of the required constraints. You should prioritize requirement satisfaction rather than relying only on general impressions such as fluency, coherence, or helpfulness. Finally, provide a ranking of the candidate responses from best to worst.

%% file: tables/constraint_prompt.tex
\begin{table*}[htbp]
\begin{tcolorbox}[colback=white, colframe=black, boxrule=1pt, width=\textwidth]
\footnotesize
\textbf{[Task Description]}

\textbf{1.} You will receive a [Seed Question]. Your task is to design five new atomic constraints that significantly increase the difficulty of answering the question.

\textbf{2.} These constraints must be added without modifying the original [Seed Question].

\textbf{3.} Constraints must be specific and granular, avoiding vague instructions (e.g., "brief", "formal"), and must not conflict with each other.

\textbf{4.} The five constraints should cover multiple types and cannot all belong to the same type.

\textbf{5.} Each constraint must include clear, actionable details such as specific numbers, scenarios, times, entities, sequences, or language structures.

\textbf{6.} Encourage the inclusion of reasoning, metaphor, classification, ordering, or background elaboration to increase difficulty.

\textbf{7.} The output must be in strict JSON format with keys c1, t1, c2, t2, c3, t3, c4, t4, c5, t5 representing five constraints and their types.

\textbf{[Constraint References]}

\textbf{1. Length Constraints}

\quad\textbf{Word Limit:} Set a word count range, e.g., "Use 120-150 words."

\quad\textbf{Sentence Limit:} Restrict sentences, e.g., "Write exactly 5 sentences."

\quad\textbf{Paragraph Limit:} Limit paragraphs and length, e.g., "Use 3 paragraphs, each at most 40 words."

\quad\textbf{Proportional Distribution:} Allocate proportions, e.g., "Conclusion is 30\% of text."

\quad\textbf{Character Limit:} Restrict characters, e.g., "at most 500 characters, including spaces."

\quad\textbf{Sentence Length:} Limit sentence length, e.g., "Each sentence less than 15 words" or "at least 10 English words."

\textbf{2. Format Constraints}

\quad \textbf{Customized formatting for specific needs}, e.g., "Summarize main points in an unordered list."

\quad \textbf{Formatting standards for specialized applications}, e.g., "Conform to electronic medical record format."

\quad \textbf{Defines how to highlight or emphasize parts of the text using styles or symbols}, e.g., "Bold all key terms and use warning symbols before warnings."

\textbf{3. Style Constraints}

\quad\textbf{Tone:} Adopt a tone, e.g., "Humorous and sarcastic."

\quad\textbf{Rhetorical Devices:} Include devices, e.g., "Use at least 2 metaphors."

\quad\textbf{Audience:} Target audience, e.g., "Explain for elementary students."

\quad\textbf{Identity/Voice:} Write from perspective, e.g., "As a historian" or "Retired elder."

\quad\textbf{Emotional Appeal:} Use emotional tone, e.g., "Evoke empathy" or "Create urgency."

\quad\textbf{Literary Device:} Include techniques, e.g., "Use 1 parallelism, 1 rhetorical question."

\quad\textbf{Cultural Voice:} Adopt cultural style, e.g., "Tang dynasty poet" or "Japanese haiku style."

\textbf{4. Content Constraints}

\quad\textbf{Required Entities:} Include entities, e.g., "Name 3 scientists."

\quad\textbf{Chronological Order:} Present sequentially, e.g., "Events in chronological order."

\quad\textbf{Data Requirement:} Include data, e.g., "Use at least 2 statistical data points."

\quad\textbf{Thematic Coverage:} Cover themes, e.g., "Economic, environmental, social aspects."

\quad\textbf{Time-Space Perspective:} Use unique perspective, e.g., "Song dynasty" or "Arctic explorer."

\quad\textbf{Interdisciplinary Focus:} Combine disciplines, e.g., "Physics and philosophy."

\quad\textbf{Counterargument:} Address opposition, e.g., "Discuss 1 counterargument, rebut with evidence."

\quad\textbf{Hypothetical Scenario:} Use hypotheticals, e.g., "Write from lunar base about future tech."

\textbf{[Seed Question]}

\{raw\_question\}
\end{tcolorbox}
\caption{Complete prompt for constructing evaluation datasets}
\label{tab:constraint_prompt}
\end{table*}

%% file: tables/eval.tex
\begin{table*}[t]
\resizebox{\linewidth}{!}{
\begin{tcolorbox}
\small

Act as an impartial judge and evaluate the quality of the written responses provided by two AI assistants to the user's writing prompt below.

You will be given Assistant A's response and Assistant B's response. Your job is to determine which assistant's writing is superior.

\textbf{Evaluation Criteria:}

\textbf{1. Relevance and Completeness:}

- Does the assistant fully respond to the writing prompt?

- Does the length meet the user's query expectations?

- Is the content relevant to the topic?

- Does it provide sufficient depth, length, and detail, rather than drifting off-topic or being simplistic?

\textbf{2. Writing Quality:}

- Evaluate whether the assistant's writing is clear, fluent, and free of obvious grammatical errors.

- The overall quality of the writing should be high, with elegance.

\textbf{3. Creativity and Originality:}

- If applicable, assess the creativity of the response.

- Does the assistant offer fresh perspectives, unique insights, or demonstrate a certain level of originality?

\textbf{4. Specificity and Detail:}

- Determine whether the assistant provides concrete examples or detailed explanations.

- Properly justified repetition is permissible.

\textbf{5. Tone and Style:}

- Is the tone appropriate for the writing prompt?

- Is the writing style consistent throughout?

- Consider whether it aligns with the expectations of the intended audience or writing purpose.

After evaluating each response based on these factors, determine which one is superior, provide an explanation, and then select one of the following final verdicts:

- \textbf{Assistant A is significantly better:} [[A»B]]

- \textbf{Assistant A is slightly better:} [[A>B]]

- \textbf{Tie, relatively the same:} [[A=B]]

- \textbf{Assistant B is slightly better:} [[B>A]]

- \textbf{Assistant B is significantly better:} [[B»A]]

Example output: My final verdict is tie: [[A=B]].

\end{tcolorbox}
}
\caption{Prompts used for the Arena-Write evaluation~\cite{wu2025longwriter}.}
\label{tab:eval}
\end{table*}